\pdfoutput=1

\documentclass[11pt]{article}

\usepackage[]{acl}

\usepackage{times}
\usepackage{latexsym}
\usepackage{soul}
\usepackage{booktabs}
\usepackage{float}
\usepackage{graphicx}
\usepackage{multirow}

\newcommand{\modelname}{MixQG}
\newcommand{\githublink}{\url{https://github.com/salesforce/QGen}}

\graphicspath{ {./fig/} }

\usepackage[T1]{fontenc}

\usepackage[utf8]{inputenc}

\usepackage{microtype}

\title{{\modelname}: Neural Question Generation with Mixed Answer Types}

\author{
  \quad \textbf{Lidiya Murakhovs'ka}
  \quad \textbf{Chien-Sheng Wu}
  \quad \textbf{Philippe Laban} \\
  \quad \textbf{Tong Niu}
  \quad \textbf{Wenhao Liu}
  \quad \textbf{Caiming Xiong} \\
  Salesforce AI Research \\
  \{l.murakhovska, wu.jason, plaban, tniu, wenhao.liu, cxiong\}@salesforce.com
}

\begin{document}
\maketitle
\begin{abstract}

Asking good questions is an essential ability for both human and machine intelligence. However, existing neural question generation approaches mainly focus on short factoid type of answers. In this paper, we introduce a neural question generator, {\modelname}, to bridge this gap. We combine nine question answering datasets with diverse answer types, including yes/no, multiple-choice, extractive, and abstractive answers, to train a single generative model. We show with empirical results that our model outperforms existing work in both seen and unseen domains, and can generate questions with different cognitive levels when conditioned on different answer types. We run a human evaluation study to assess the quality of generated questions and find that {\modelname} outperforms the next best model by 10\%. Our code and model checkpoints will be released and integrated with the HuggingFace library to facilitate various downstream applications.

\end{abstract}

\section{Introduction}

Question generation (QG) aims to automatically create questions from a given text passage or document with or without answers. It has a wide range of applications such as improving question answering (QA) systems~\cite{duan-etal-2017-question} and search engines~\cite{10.1145/3308558.3313746} through data augmentation, making chatbots more engaging~\cite{wang-etal-2018-learning-ask, laban2020s}, enabling automatic evaluation~\cite{rebuffel2021data} and fact verification~\cite{pan-etal-2021-zero}, and facilitating educational applications~\cite{chen2018learningq}.

\begin{figure}[t!]
\normalsize
  \center{
  \resizebox{\linewidth}{!}{
  \small
  \setlength{\tabcolsep}{0.0in}
    \begin{tabular}{p{7.7cm}}
    \hline 
    \textbf{Context}: In the late 17th century, Robert Boyle proved that air is necessary for combustion. English chemist John Mayow (1641–1679) refined this work by showing that fire requires only a part of air that he called spiritus nitroaereus or just nitroaereus. In one experiment he found that placing either a mouse or a lit candle in a closed container over water caused the water to rise and replace one-fourteenth of the air's volume before extinguishing the subjects. From this he surmised that nitroaereus is consumed in both respiration and combustion.\\
    \hline
    \textbf{Question}: Who proved that air is necessary for combustion? \\
    \textbf{Ext. Short Answer}: Robert Boyle\\ 
    \hline
    \textbf{Question}: How did John Mayow find that spiritus nitroaereus is consumed in both respiration and combustion? \\
    \textbf{Abs. Short Answer}: through an experiment\\
    \hline
    \textbf{Question}: Does fire need air to burn?        \\   
    \textbf{Yes/No Answer}: yes\\ 
    \hline
    \textbf{Question}: What did John Mayow discover about nitroaereus? \\
    \textbf{Ext. Long Answer}: In the late 17th century … in both respiration and combustion. \\  
    \hline
    \textbf{Question}: Why was the mouse used in the experiment? \\
    \textbf{Abs. Long Answer}: The mouse was used in the experiment to test the consumption of nitroaereus during respiration. \\    
    \hline
    \end{tabular}     
  }
  }
  \caption{Given the same context, {\modelname} generates diverse questions based on the target answer choice.}
\label{figure:qa}
\end{figure}

\begin{table*}[t]
\centering
\resizebox{\linewidth}{!}{
\begin{tabular}{@{}lllll@{}}
\toprule
\textbf{Dataset} & \textbf{Type}   & \textbf{Source}               & \textbf{Train examples} & \textbf{Dev. examples} \\ \midrule
SQuAD~\cite{rajpurkar-etal-2016-squad}            & Extractive     & Wikipedia                     & 86,588                  & 10,507                \\
NewsQA~\cite{trischler-etal-2017-newsqa}           & Extractive      & News                          & 74,160                  & 4,212                 \\
TriviaQA~\cite{joshi-etal-2017-triviaqa}         & Extractive      & Web                           & 61,688                  & 7,785                 \\
SearchQA~\cite{DBLP:journals/corr/DunnSHGCC17}         & Extractive      & Web                           & 117,384                 & 16,980                \\
HotpotQA~\cite{yang-etal-2018-hotpotqa}         &  Extractive     & Wikipedia                     & 72,928                  & 5,904                 \\
NQ~\cite{kwiatkowski-etal-2019-natural}               & Extractive      & Wikipedia                     & 104,071                 & 12,836                \\
NarQA~\cite{kocisky-etal-2018-narrativeqa}            &  Abstractive    & Wikipedia,  Project Gutenberg & 32,747                  & 3,461                 \\
MCTest~\cite{richardson-etal-2013-mctest}           & Multiple-Choice      & Stories                       & 1,200                   & 600                   \\
BoolQ~\cite{clark-etal-2019-boolq}            & Yes-No      & Wikipedia                     & 9,427                   & 3,270                 \\ \hline
Quoref*~\cite{Dasigi2019Quoref}           & Extractive       & Wikipedia             & 19,399 & 2,418 \\ 
QAConv*~\cite{wu2021qaconv}           & Extractive      & Email, Panel, Channel & 25,988 & 3,251 \\ 
DROP*~\cite{Dua2019DROP}             & Abstractive     & Wikipedia             & 77,400 & 9,535 \\
TweetQA*~\cite{xiong-etal-2019-tweetqa}             & Abstractive     & Twitter             & 10,692 & 1,086 \\
\bottomrule
\end{tabular}
}
\caption{Dataset Statistics of various QA corpora. * indicates unseen corpus during training.}
\label{table:datasetStatistics}
\end{table*}

Earlier QG approaches relied on syntactic rules that incorporated linguistic features into the QG process~\cite{heilman-smith-2010-good, khullar-etal-2018-automatic}. \citet{DBLP:journals/corr/DuSC17} pointed out some of the limitations of such rule-based systems and formulated the task of question generation as a sequence-to-sequence learning problem. Based on this formulation, recent works rely on pre-trained Transformer-based models to generate answer-aware questions~\cite{DBLP:journals/corr/abs-1905-03197,yan2020prophetnet,newsquiz2021}. However, the majority of QG research so far has been performed on the SQuAD dataset~\cite{rajpurkar-etal-2016-squad}, and as a result, it mainly focuses on factoid short answer questions~\cite{zhang-bansal-2019-addressing, zhou-etal-2019-question, su-etal-2020-multi}. 

In reality, answers can come in a variety of types and forms, e.g., short/long, multiple-choice, yes-no, and extractive/abstractive answers. We hypothesize that \textit{answer types are as important as question types}, and that different answer types have their unique QG challenges and result in questions with different cognitive levels. {\modelname} combines nine QA datasets with varied answer types to build a more robust and versatile QG model. We use pre-trained generative language models like T5~\citep{2020t5} and BART~\citep{DBLP:journals/corr/abs-1910-13461} without question-specific or domain-specific prefixes to generate the questions. Figure~\ref{figure:qa} illustrates the above, showing {\modelname}-generated questions of different cognitive levels for different answer types.

The contribution of this paper is summarized as follows:
1) We train a unified QG model that achieves state-of-the-art performance in both seen and unseen domains. We release training code and model checkpoints (base, large, 3B) to facilitate various downstream QG applications \footnote{{\githublink}}.
2) We show that {\modelname} is able to produce different cognitive level questions by controlling the answer types. We conduct a human evaluation study which confirms that {\modelname} leads to improvements in question quality in a practical quiz design setting.

\section{Methodology}

\subsection{Datasets}
\label{sec:Dataset}
We leverage nine commonly used QA datasets (Table~\ref{table:datasetStatistics}) to train our {\modelname} model, including six MRQA 2019 Shared Task~\cite{DBLP:journals/corr/abs-1910-09753} datasets, NarrativeQA~\cite{kocisky-etal-2018-narrativeqa}, MCTest~\cite{richardson-etal-2013-mctest}, and BoolQ~\cite{clark-etal-2019-boolq}. These represent the majority of large-scale publicly available QA datasets. We obtain in total 560,193 training examples with different answer types and source domains. We reserve their validation set for in-domain evaluation.

In most general sense, a QA dataset comprises of $<$C, Q, A$>$ tuples, where $C$ is a context document, $Q$ is a human-written question, and $A$ is its corresponding answer.
Following a common classification of answer types, we bucket each dataset into one of the below categories: 1) \textbf{Extractive [EX]}: the answer to the question is a substring of the context passage. 2) \textbf{Abstractive [AB]}: the answer to the question is written in free-form and is not necessarily contained within the context passage. 3) \textbf{Multiple-Choice [MC]}: question comes with multiple answers to select from, including a single correct option and several distractors. 4) \textbf{Yes-No [YN]}: the answer is a boolean response. Datasets that do not comply with the above format, such as ELI5~\cite{fan-etal-2019-eli5} and GooAQ~\cite{khashabi-etal-2021-gooaq-open}, were excluded from training. We leave their exploration to future work.

We also leverage a set of datasets unseen during training to evaluate our model’s generalization ability. Similar to the train datasets, these cover several text sources, domains, and answer types. Quoref~\cite{Dasigi2019Quoref} questions can have disjoint spans as answers and often require coreference resolution. DROP~\cite{Dua2019DROP} questions require discrete reasoning over the context paragraphs. QAConv~\cite{wu2021qaconv} uses informative conversations such as emails, channels, and panels as a knowledge source, and it includes extractive answers from multiple text spans. TweetQA~\cite{xiong-etal-2019-tweetqa} uses social media as an information source and contains abstractive answers.

Note that to generate fluent questions, we need to place some restrictions on the training data we use. For example, we disregard "fill-in-the-blank" (a.k.a Cloze-style) reading comprehension datasets as their questions are implicit and thus do not aid the QG model. Similarly, we ensure that our training data does not contain unanswerable questions or multiple-choice questions that are too general (e.g., ``which of the following is TRUE according to the passage?'').

\begin{table}[H]
\fontsize{10pt}{10pt}\selectfont
\centering
\begin{tabular}{@{}ll@{}}
\toprule
\textbf{Type} & \textbf{Input}                                           \\ \midrule
EX            & \{answer\} \textbackslash{}n \{context\}                        \\
AB            & \{answer\} \textbackslash{}n \{context\}                        \\
MC            & \{correct\_answer\} \textbackslash{}n \{context\}               \\
YN            & \{answer\} + \{entities\} \textbackslash{}n \{context\}         \\ \bottomrule
\end{tabular}
\caption{Input answer formatting.} 
\vspace{-3mm}
\label{table:formatting}
\end{table}

\subsection{Language Modeling}

\begin{table*}[!htbp]
\centering
\resizebox{0.85\linewidth}{!}{
\begin{tabular}{@{}lllccccccc@{}}
\toprule
\textbf{Dataset}     & \textbf{Model}     &  \textbf{Size}        & \multicolumn{1}{l}{\textbf{BLEU}} & \multicolumn{1}{l}{\textbf{R1}} & \multicolumn{1}{l}{\textbf{R2}} & \multicolumn{1}{l}{\textbf{RL}} & \multicolumn{1}{l}{\textbf{RLsum}} & \multicolumn{1}{l}{\textbf{METEOR}} & \multicolumn{1}{l}{\textbf{BERTScore}} \\ \midrule
            & ProphetNet-pre & large
            & 22.88                  & 51.37                    & 29.48                    & 47.11                    & 47.09                       & 41.46                     & 0.4931                        \\
            & BART-hl  & base      & 21.13                  & 51.88                     & 29.43                    & 48.00                    & 48.01                       & 40.23                     & 0.5433                        \\
            & T5-hl    & base      & 23.19                  & 53.52                    & 31.22                    & 49.40                    & 49.40                       & 42.68                     & 0.5548                        \\
SQuAD       & BART-pre  & base   & 22.09    & 52.75     & 30.56    & 48.79   & 48.78   & 41.39    & 0.5486 \\
            & T5-pre    & base       & \textbf{23.74}               & 54.12                       & 31.84           & 49.82   & 49.81 & 43.63 & 0.5568 \\
            & {\modelname}     & base         & 23.53                  & 54.39                    & 32.06                   & 50.05                    & 50.02                       & 43.83                     & 0.5566                        \\
            & {\modelname}$_{finetuned}$   & base     & 23.46         & \textbf{54.48}           & \textbf{32.18}           & \textbf{50.14}           & \textbf{50.10}               & \textbf{44.15}            & \textbf{0.5582}               \\
            \cline{2-10}
            & {\modelname} & 3B  & \textbf{25.42}	& \textbf{56.11}	& \textbf{33.91}	& \textbf{51.85}	& \textbf{51.86}	& \textbf{45.75}	& \textbf{0.5789} \\
            \hline
            & T5-pre     & base         & 29.99                  & 59.53                    & 37.83                    & 56.65                    & 56.64                       & 54.38                     & 0.5202                        \\
NQ   & {\modelname}   & base           & 30.69                  & 60.04                     & 38.43                    & 57.09                   & 57.09                        & 54.76                     & 0.5246                        \\
            & {\modelname}$_{finetuned}$   & base        & \textbf{31.25}         & \textbf{60.98}            & \textbf{39.21}           & \textbf{57.84}           & \textbf{57.84}              & \textbf{55.90}             & \textbf{0.5351}               \\ 
            \cline{2-10}
            & {\modelname} & 3B  & \textbf{33.91} &	\textbf{63.17} &	\textbf{41.95} &	\textbf{60.15} &	\textbf{60.15} &	\textbf{58.34} &	\textbf{0.5610} \\
            
\bottomrule

\end{tabular}
}
\caption{Results on two seen datasets, SQuAD~\cite{rajpurkar-etal-2016-squad} and NQ~\cite{kwiatkowski-etal-2019-natural}.}
\label{table:joint}
\end{table*}

\begin{table*}[!htbp]
\centering
\resizebox{0.85\linewidth}{!}{
\begin{tabular}{@{}lllccccccc@{}}
\toprule
\textbf{Answer Type} & \textbf{Dataset}     &  \textbf{Model}        & \multicolumn{1}{l}{\textbf{BLEU}} & \multicolumn{1}{l}{\textbf{R1}} & \multicolumn{1}{l}{\textbf{R2}} & \multicolumn{1}{l}{\textbf{RL}} & \multicolumn{1}{l}{\textbf{RLsum}} & \multicolumn{1}{l}{\textbf{METEOR}} & \multicolumn{1}{l}{\textbf{BERTScore}} \\ \midrule

          &             & T5-pre            & 21.32                  & 45.94                     & 27.91                    & 42.92                    & 42.90                       & 38.27                     & 0.4374                        \\
EX & QAConv                 & {\modelname}           & 16.65                  & 39.99                    & 22.01                    & 37.62                    & 37.59                       & 29.07                     & 0.4117                        \\
           &             & {\modelname}$_{finetuned}$     & \textbf{22.74}         & \textbf{47.40}           & \textbf{29.48}           & \textbf{44.41}           & \textbf{44.40}              & \textbf{39.93}            & \textbf{0.4533}               \\
\hline
           &            & T5-pre            & 26.88                  & 45.54                    & 31.98                    & 44.10                    & 44.12                       & 41.84                     & \textbf{0.4150}                         \\
EX & Quoref                 & {\modelname}        & 4.28                   & 24.89                    & 7.97                     & 22.27                     & 22.30                       & 14.13                     & 0.2859                        \\
            &           & {\modelname}$_{finetuned}$ & \textbf{27.36}         & \textbf{45.91}           & \textbf{32.41}           & \textbf{44.42}           & \textbf{44.42}              & \textbf{42.06}            & 0.4137               \\
\hline
            &           & T5-pre         & 28.46                  & 53.48                     & 35.49                     & 50.97                    & 51.00                        & 47.50                      & 0.5491                        \\
AB & DROP                   & {\modelname}       & 7.16                   & 30.66                    & 12.95                    & 28.38                    & 28.40                       & 23.23                     & 0.3556                        \\
            &            & {\modelname}$_{finetuned}$    & \textbf{28.53}         & \textbf{53.72}           & \textbf{35.63}           & \textbf{51.11}           & \textbf{51.12}               & \textbf{47.83}            & \textbf{0.5493} \\
\hline
            &            & T5-pre           & 17.02                  & 45.28                     & 23.28                     & 44.20                    & 44.18                        & 44.63                    & 0.4384                        \\
AB & TweetQA                   & {\modelname}       & 5.28                   & 28.18                    & 10.65                    & 26.91                    & 26.89                       & 28.83                     & 0.2653                        \\
            &            & {\modelname}$_{finetuned}$    & \textbf{18.66}         & \textbf{47.12}           & \textbf{24.95}           & \textbf{45.97}           & \textbf{45.94}               & \textbf{46.60}            & \textbf{0.4645} \\

\bottomrule

\end{tabular}
}
\caption{Results on unseen datasets, QAConv \cite{wu2021qaconv}, Quoref \cite{Dasigi2019Quoref}, DROP \cite{Dua2019DROP}, and TweetQA \cite{xiong-etal-2019-tweetqa}. All models are of size base.}
\label{table:joint_unseen}
\end{table*}

We rely on a text-to-text framework as a basis for {\modelname} (Training details are in Section \ref{sec:training_details}). When combining our training datasets, we encode all inputs and outputs into a unified plain-text format. For answer-aware question generation, the input is usually formatted in one of the two ways: (1) prepending (\textbf{-pre}) the answer before the context and separating it from the rest of the text by a special separator token or (2) highlighting (\textbf{-hl}) the answer span within the context with special highlight tokens ~\citep{chan-fan-2019-recurrent}. To maintain flexibility, we rely on prepending the answer since highlighting is only applicable to the extractive answer types. In particular, we format the inputs to our model such that the answer always precedes the context paragraph and use a ``\textbackslash n'' separator in between, as shown in Table \ref{table:formatting}.

For MC type of data, we only take the correct answer and disregard the distractor options. For YN data, we extract entities from the question using spaCy's NER model \footnote{https://spacy.io/api/entityrecognizer} and append them to the answer. The reason for adding additional entities is to restrict the domain of questions, as given a context paragraph, there are many boolean questions whose answer would be yes or no, without further restriction. 
Note that no type-specific prefixes are added to the input representation, and the corresponding questions are used as output. 

\section{Experimental Results}

\subsection{Automatic Metrics}
We report the commonly-used metrics applied in the QG research: BLEU~\cite{papineni-etal-2002-bleu}, ROUGE~\cite{lin-2004-rouge}, and METEOR~\cite{banerjee-lavie-2005-meteor} scores. We also report BERTScore~\cite{DBLP:conf/iclr/ZhangKWWA20}, which relies on contextual embeddings to produce the final score.

\subsection{In-Domain Analysis}

In Table~\ref{table:joint}, we compare baselines trained solely on the target in-domain dataset against {\modelname} and {\modelname}$_{finetuned}$. {\modelname} indicates our model that is joint trained on nine QA datasets with random sampling, and {\modelname}$_{finetuned}$ is the one further fine-tuned on the target dataset. 
We show results on two datasets: SQuAD and NQ. Since SQuAD is the most common benchmark for QG, we additionally compare {\modelname} against existing question generation models such as ProphetNet~\cite{qi-etal-2020-prophetnet} and other T5 variants.
The results show that {\modelname} outperforms an equally sized model trained directly on the target dataset. Given that question styles and dataset domains may vary across {\modelname}'s seed datasets, additional fine-tuning on the target dataset further improves the scores. This shows that {\modelname} is a strong pretrained model which can be further adapted to specific use cases.

\subsection{Out-of-Domain Analysis}

Table \ref{table:joint_unseen} summarizes the evaluations on out-of-domain datasets of extractive and abstractive answer types. We observe that a dedicated model trained on the target dataset outperforms {\modelname} in a zero-shot setting. 
One potential reason is that answer and question style in different QA datasets may differ significantly.
For example, answers are ambiguous pronouns in the Quoref dataset, and questions in DROP dataset are intentionally created for discrete reasoning.
However, {\modelname}$_{finetuned}$ obtains the best overall scores after further fine-tuning on the target training set, suggesting that {\modelname} is a strong starting point for further fine-tuning question generation models.

\subsection{Human Evaluation}
\label{sec:human_eval}
Recent studies have shown that n-gram based metrics may not correlate well with human judgements \citet{nema2018towards}. The objective of human evaluation is to evaluate QG models by measuring how useful they are as a tool to aid teachers in quiz creation. 
We compare seven QG models and collect 3,164 human-annotated samples from 10 recruited teachers. More details are in Section \ref{sec:appn_quiz}.

\paragraph{Quiz Design Task}

Given an article on the quiz topic selected from Wikipedia, teachers are asked to specify a quiz concept (a subset of the article) they want to test their students on. This is used as the target answer input for QG models. Teachers can then approve a generated question to be included on the quiz or reject it and provide a reason for rejection.
The success of a QG model depends on its question approval rate.

\begin{figure}
    \centering
    \includegraphics[width=0.49\textwidth]{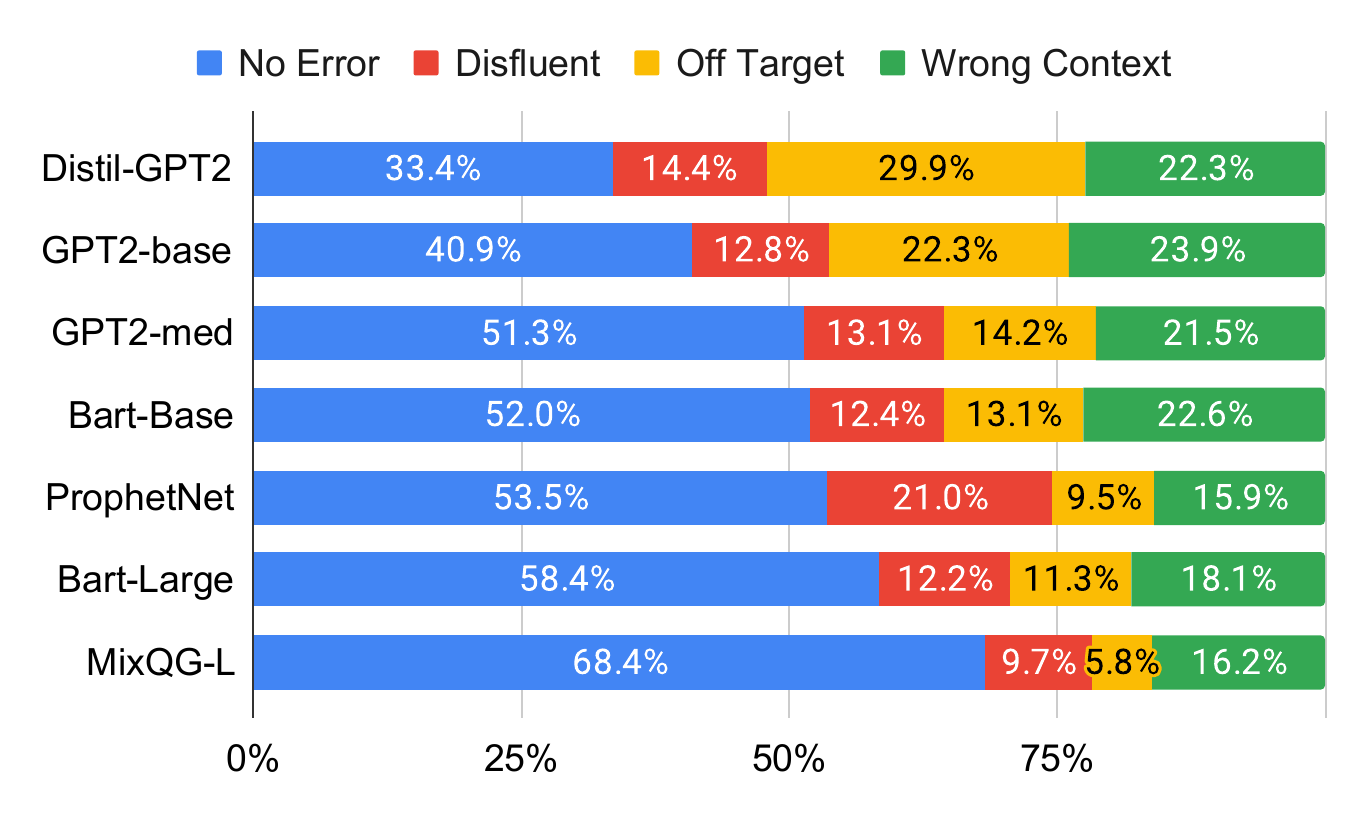}
    \caption{Human approval rate of seven QG models.}
    \label{fig:error_distribution}
\end{figure}

Besides {\modelname}, three GPT2 baselines~\cite{radford2019language}, two BART baselines~\cite{DBLP:journals/corr/abs-1910-13461}, and ProphetNet-Large finetuned on SQuAD are evaluated. In Figure \ref{fig:error_distribution}, we see that {\modelname} attains a 68.4\% acceptance rate, outperforming the next best model by 10\%. 
{\modelname} also generates the smallest number of disfluent and off target (answer mismatch) questions - with majority of errors coming from wrong context (too general or too specific) questions. Generating questions with the right level of specificity remains a challenge and is a promising direction for future work.

\subsection{Qualitative Analysis}
First, we compare {\modelname} generated questions to the gold questions annotated in five public QA datasets (Table \ref{table:quality_ex}). 
We find that the generated questions are fluent, relevant, and reasonable to the provided answer and context, even if they differ from the gold label.
This further motivates the need of human evaluation for QG research.

Second, we use the HuggingFace summarization pipeline to obtain the summary of the context, and we feed each sentence of the summary as the target answer to {\modelname} to obtain questions. In this way, we can test {\modelname}'s generalization ability to abstractive answers. As shown in Figure \ref{fig:quality}, we observe that feeding in long and abstractive answers can still generate fluent and reasonable questions, suggesting that it is possible to control the question's cognitive level by its answer. We leave as future work further research into summary-based unsupervised QA-pair generation.

Lastly, in the Quiz Design study, we find there are 106 cases in which the teachers only accepted a single candidate question into the quiz. {\modelname} produced the accepted candidate 47 times, more than any of the other models. 
We provide three examples of such {\modelname}-only success cases as well as three instances in which the {\modelname}'s question was not accepted in Table \ref{table:mixqg_study_examples}.

\section{Related Work}

Question generation's practical importance has lead to an increasing interest in the field. The early work in QG relied on linguistic templates and rules to produce questions from declarative sentences ~\citep{heilman-smith-2010-good, labutov-etal-2015-deep}. With the success of neural techniques in text generation tasks, applying neural sequence-to-sequence generation models became more common ~\citep{DBLP:journals/corr/DuSC17, sun-etal-2018-answer}. More recent works leverage pre-trained transformer based networks, such as T5~\citep{2020t5}, BART~\citep{DBLP:journals/corr/abs-1910-13461}, PEGASUS~\cite{DBLP:journals/corr/abs-1912-08777} and ProphetNet~\cite{DBLP:journals/corr/abs-2001-04063}, for question generation which have been successful in many applications ~\citep{unilm, newsquiz2021, rebuffel2021data, pan-etal-2021-zero}.

However, most of the earlier work focuses on using a single QA dataset, such as SQuAD~\cite{rajpurkar-etal-2016-squad}. While working on generation of open-ended~\cite{DBLP:journals/corr/abs-2107-00152}, controllable~\cite{DBLP:journals/corr/abs-2107-00152}, multi-hop~\cite{cho-etal-2021-contrastive} or cause-effect~\cite{stasaski-etal-2021-automatically} questions has gained attention, each direction is studied in isolation as it usually requires a separate QA dataset.

Most directly related to our work is UnifiedQA~\cite{khashabi-etal-2020-unifiedqa}, which successfully crosses format boundaries of different QA datasets to train a robust QA system. It advocates for more general and broader system designs not limited to specific dataset formats. Similar to their approach, {\modelname} combines multiple QA datasets and trains a single QG system in a text-to-text paradigm.

\section{Conclusion}

In this paper, we present {\modelname}, a question generation model pre-trained on a collection of QA datasets with a mix of answer types. We show through experiments that the resulting model is a strong starting point for further fine-tuning which achieves state-of-the-art results on target datasets in commonly-used similarity metrics as well as our designed human evaluation. We release our code and the model checkpoints to facilitate QG research and downstream applications.

\section{Ethical Considerations}
{\modelname} is subject to biases found in the training data of both the underlying text-to-text models and all QA datasets that we have used for pre-training. We do not collect a new dataset for question generation and instead reuse data from previously published works. As such, we rely on the published works to follow the responsible data collection practices. The model is currently English language only which limits its practical applications in the real world. We hope to make {\modelname} multilingual as more diverse QA datasets become available in the future. We validate the proposed model by conducting a human evaluation. We recruited 10 teachers for a study that lasted a maximum of two hours and gifted each participant a \$50 gift card.

\bibliographystyle{acl_natbib}
\bibliography{anthology,main}

\appendix

\section{Training Details}
\label{sec:training_details}
Training datasets are listed in Table \ref{table:datasetStatistics}. For training {\modelname}, we use several pre-trained text-to-text model checkpoints from the HuggingFace library~\cite{wolf-etal-2020-transformers}. We finetune them for question generation using our combined dataset described in Section \ref{sec:Dataset}. For most experiments done in this paper, we finetune on a T5-base model~\cite{2020t5}. We also scale up the model and report results for T5-large, T5-3B, and BART-large settings (Appendix \ref{app:scaling}). We train for 100,000 steps (or 22 epochs) with a learning rate of $3\times10^{-5}$ using the AdamW~\cite{DBLP:journals/corr/abs-1711-05101} optimizer and a batch size of 32. All training was done on eight A100 NVIDIA GPUs and took approximately 35 hours.

\section{Quiz Design Task Details}
\label{sec:appn_quiz}
We recruit teachers or ex-teachers from an online group forum. In total, 20 participants filled out the interest form, 14 were selected, and 10 completed the study. The participants had been teachers for at least a year and 3.6 years on average, and had taught diverse subjects such as sciences, history, literature, and IT topics, at various levels from primary school to college-level. The study was meant to last a maximum of two hours, and participants were gifted a \$50 gift card upon completion.

\begin{figure}
    \centering
    \includegraphics[width=0.44\textwidth]{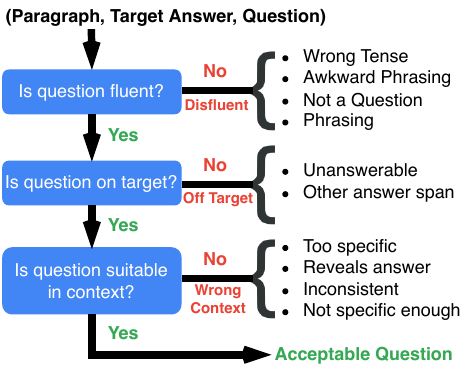}
    \caption{Hierarchical categorization of errors for question generation. Three error categories (Disfluent, Off Target, Wrong Context) each with several subtypes.}
    \label{fig:qgen_error_typology}
\end{figure}

\begin{figure*}
    \centering
    \frame{\includegraphics[width=0.95\textwidth]{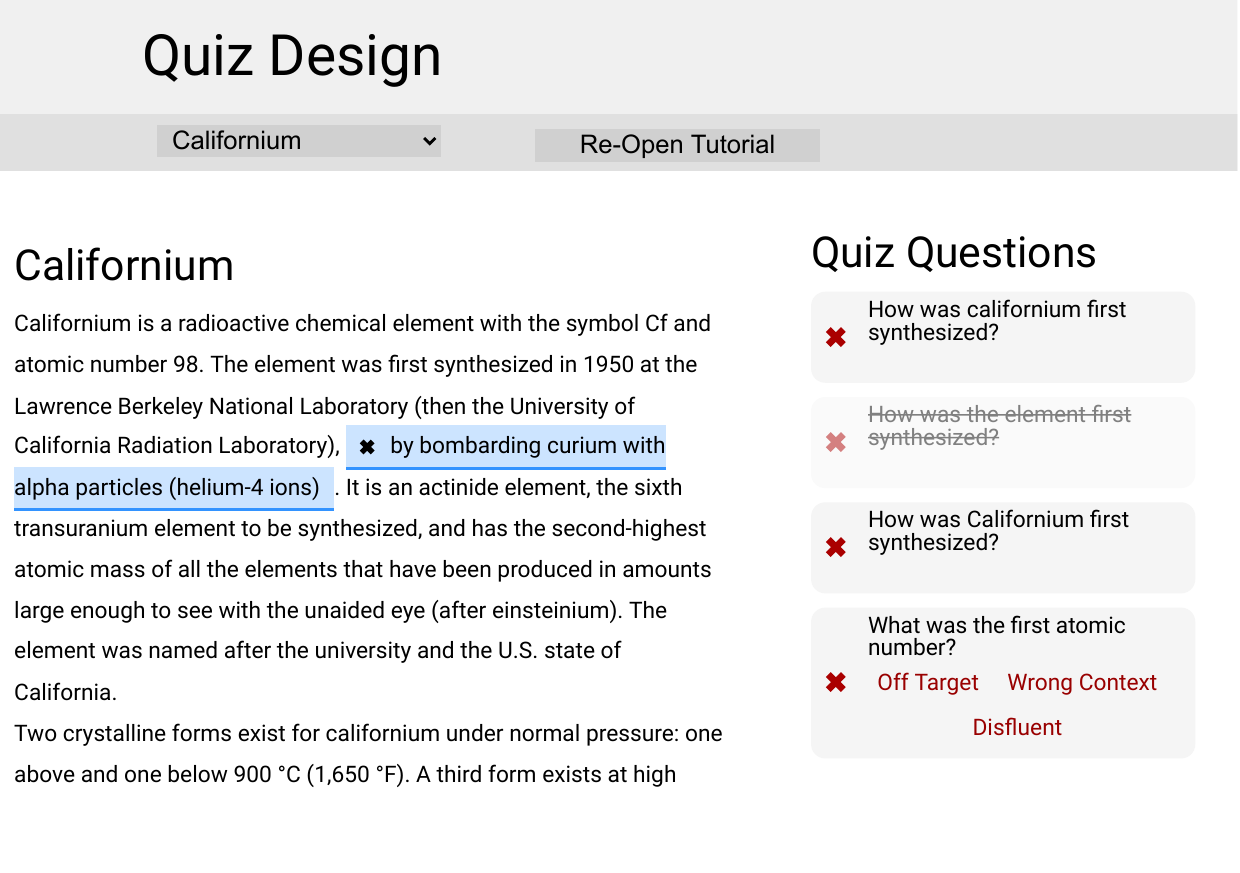}}
    \caption{\textbf{Screenshot of annotation interface used for the Quiz Design Task.} The teacher has selected the concept highlighted in blue in the reading material in the left column. In the right column, the system gives proposes candidate questions, which can be added to the quiz, or refused with a reason.}
    \label{fig:quiz_design_interface}
\end{figure*}

Participants were tasked with creating between 5-7 quizzes, each with a minimum of 8 concepts, and could pick from a set list of 7 quiz topics, which we pre-selected from the list of featured Wikipedia articles\footnote{\url{https://en.wikipedia.org/wiki/Wikipedia:Featured_articles}}. We purposefully selected articles within different domains to benchmark the QGen models in diverse topical settings: two in physics (Sustainable Energy, Californium Atom), two in biology (DNA, Enzymes), two in history (Statue of Liberty, Palazzo Pitti), and one in geology (the K-T extinction). Participants were given the first 500 words of the Wikipedia page of each topic as reading material to select Quiz concepts from. User interface is shown in Figure \ref{fig:quiz_design_interface}. Hierarchical categorization of errors for question generation is shown in Figure \ref{fig:qgen_error_typology}.

\section{Qualitative Study Details} \label{app:qualitative_study}
\begin{figure*}[t]
\centering
\includegraphics[width=\textwidth]{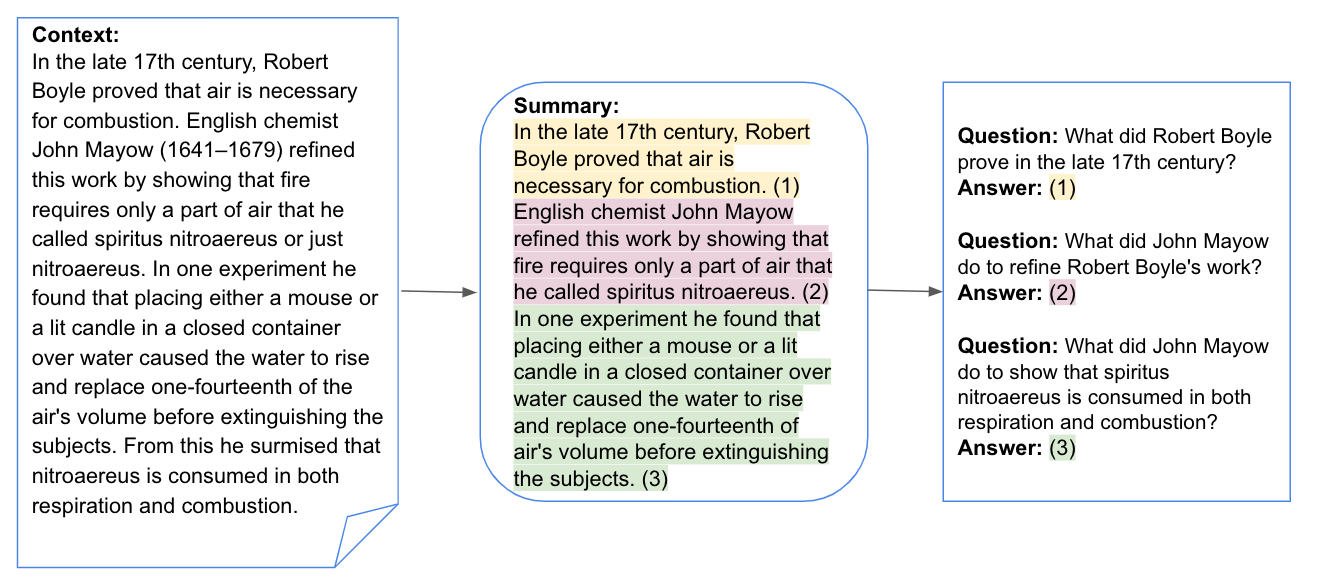}
\caption{Example of generating QA pairs using summarization and {\modelname}.}
\label{fig:quality}
\end{figure*}

To understand {\modelname}’s performance beyond automated metrics, we analyze its generated questions in Table \ref{table:quality_ex}. It shows several examples of questions generated by {\modelname}-3B on the validation sets of different datasates along with the ground-truth questions. 
We also generate question-answer pairs on Wikipedia articles using a pipeline approach as shown in Figure \ref{fig:quality}. First, we use a summarization model~\footnote{\url{https://huggingface.co/facebook/bart-large-cnn}} to obtain the summary of the context. Then we feed each sentence of the summary as the target answer to {\modelname} and obtain the questions. We observe that the generated questions are grammatically fluent, relevant to the input, and answerable by the target answer paragraph. We find that feeding in longer answers to the model generates more general, higher-level questions about the source article, while short answers prompt more factoid-style questions. As a result, we are able to generate questions of varied cognitive levels from the same source document by restricting the answer part of the input. 

\section{Scaling} \label{app:scaling}

Table \ref{table:scaled} shows the performance of differently sized {\modelname} models on SQuAD dataset. We additionally train {\modelname} model based on BART-large checkpoint, referred to as {\modelname}$^{BART}_{large}$. As expected, the largest {\modelname} model (3 billion parameters) performs best among the different model size variants.

\onecolumn

\begin{table*}[!htbp]
\centering
\fontsize{13pt}{13pt}\selectfont
\resizebox{\linewidth}{!}{
\begin{tabular}{@{}lll@{}}
\toprule
Dataset & Source & Questions                                                                                          \\ \midrule
SQuAD   & Gold   & What happened to NASA's yearly budget after the first landing?                                     \\
        & {\modelname}  & What happened to NASA's budget after the first successful moon landing?                            \\
        \cline{2-3}
        & Gold   & How many of Warsaw's inhabitants spoke Polish in 1933?                                             \\
        & {\modelname}  & In 1933, how many of Warsaw's inhabitants were of Polish mother tongue?                            \\
        \cline{2-3}
        & Gold   & How long does it take for new areas to have significant oil production?                            \\
        & {\modelname}  & How long did it take to develop new oil fields?                                                    \\
        \hline
NarQA   & Gold   & What are Mulder and Scully doing at the beginning of the story?                                    \\
        & {\modelname}  & What are Mulder and Scully doing in Dallas?                                                        \\
        \cline{2-3}
        & Gold   & How does Chris make extra money?                                                                   \\
        & {\modelname}  & What does Chris Hughes do for a living?                                                            \\
        \cline{2-3}
        & Gold   & Who died in this story?                                                                            \\
        & {\modelname}  & What is the name of Mortimer Tregennis's sister?                                                   \\
        \hline
MCTest  & Gold   & How many of Mikes friends came to the party                                                        \\
        & {\modelname}  & How many of Mike's friends came to the party?                                                      \\
        \cline{2-3}
        & Gold   & Where did Jenny want to go on a trip to?                                                           \\
        & {\modelname}  & Where did Jenny want to go on a trip?                                                              \\
        \cline{2-3}
        & Gold   & What was Stewart doing at the beach?                                                               \\
        & {\modelname}  & What was Stewart doing when he saw the bug?                                                        \\
        \hline
BoolQ   & Gold   & is there an amtrak station in pensacola florida                                                    \\
        & {\modelname}  & is there an amtrak station in pensacola florida                                                    \\
        \cline{2-3}
        & Gold   & can a minor sit at a bar in nj                                                                     \\
        & {\modelname}  & can a minor drink at a bar in new jersey                                                           \\
        \cline{2-3}
        & Gold   & can a polar bear and a grizzly mate                                                                \\
        & {\modelname}  & is there such a thing as a grizzly polar bear                                                      \\
        \hline
QAConv  & Gold   & What Xbox 360 game is the best dance game for a wife on the Wii?                                   \\
        & {\modelname}  & What is a good dance game for the Wii?                                                             \\
        \cline{2-3}
        & Gold   & Which state is taking its proper place and represents the country's diversity much better?         \\
        & {\modelname}  & What state is taking its proper place as early in the primary?                                     \\
        \cline{2-3}
        & Gold   & In what month will the BLM say go ahead and start the stimulation experiment ? \\
        & {\modelname}  & When will the experiment start?                                                   \\ \bottomrule
\end{tabular}
}
\caption{Example of {\modelname}-3B generated questions and their corresponding ground-truth labels.}
\label{table:quality_ex}
\end{table*}

\begin{table*}
    \centering
    \resizebox{0.98\textwidth}{!}{%
    \begin{tabular}{p{5.5cm} p{1.7cm} p{5.5cm} p{1.7cm}}
    \toprule{}
    \textbf{Context} & \textbf{Model} & \textbf{Question} & \textbf{Annotation} \\
    \midrule{}
    \multirow{2}{5cm}{For example, californium can be used {[}...{]} when studying materials using \textbf{neutron diffraction and neutron spectroscopy}.} & {\modelname}-L & What methods are used to study materials using californium as a source of neutrons? & No Error \\
     & BART-L & What is Californium used to study materials? & Disfluent \\
    \midrule{}
    \multirow{2}{5cm}{With the exception of some ectothermic species {[}...{]} no tetrapods \textbf{weighing more than 25 kilograms} (55 pounds) survived.} & {\modelname}-L & What size tetrapods did not survive the extinction? & No Error \\
     & ProphetNet & How much did tetrapods weigh at the time of the Cretaceous-Paleogene extinction? & Off Tgt. \\
    \midrule{}
    \multirow{4}{5cm}{The two DNA strands are known as polynucleotides as they are composed of \textbf{simpler monomeric units called nucleotides}.} & {\modelname}-L & What are polynucleotides composed of? & No Error \\
    & & & \\
     & BART-L & What are polynucleotides? & Off Tgt. \\
     & & & \\
    \hline \hline
    \multirow{2}{5cm}{The Statue of Liberty (\textbf{Liberty Enlightening the World}) is a colossal neoclassical sculpture on {[}...{]}} & ProphetNet & What is another name for the Statue of Liberty? & No Error \\
     & {\modelname}-L & What is the English translation of the Statue of Liberty? & Off Tgt. \\
    \midrule{}
    \multirow{2}{5cm}{Californium. The element was named \textbf{after the university and the U.S. state of California}.} & ProphetNet & What is Californium named after? & No Error \\
     & {\modelname}-L & Where did Californium get its name? & Wrong Ctxt \\
    \midrule{}
    \multirow{2}{5cm}{\textbf{Fossil fuels provide 85\% of the world's energy consumption} and the energy system {[}...{]}} & BART-L & How much of the world's energy consumption does fossil fuels provide? & No Error \\
     & {\modelname}-L & What percentage of the world's energy consumption is fossil fuels? & Disfluent \\
     \bottomrule{}
    \end{tabular}
    }
    \caption{\textbf{Success and failure cases of the {\modelname} model from the Quiz Design evaluation.} Comparisons to the ProphetNet and BART-Large models are included, with each model receiving the context with a target answer (in bold), and being annotated with an error label by a teacher.}
    \label{table:mixqg_study_examples}
\end{table*}

\begin{table*}[!htbp]
\centering
\begin{tabular}{@{}lrrrrrrr@{}}
\toprule
Model          & \multicolumn{1}{l}{BLEU} & \multicolumn{1}{l}{R1} & \multicolumn{1}{l}{R2} & \multicolumn{1}{l}{RL} & \multicolumn{1}{l}{RLsum} & \multicolumn{1}{l}{METEOR} & \multicolumn{1}{l}{BERTScore} \\
\midrule
ProphetNet$_{large}$ & 22.88                           & 51.37                             & 29.48                             & 47.11                             & 47.09                                & 41.46                              & 0.4931                                 \\
\hline
{\modelname}$^{BART}_{large}$ & 23.30	& 54.44	& 31.92	& 50.18	& 50.18	& 43.47	& 0.5622 \\
\hline
{\modelname}$_{base}$                          & 23.53                           & 54.39                             & 32.06                             & 50.05                             & 50.02                                & 43.83                              & 0.5566                                 \\
{\modelname}$_{large}$                         & 24.42                           & 55.52                             & 33.13                             & 50.99                             & 50.97                                & 45.07                              & 0.5699                                 \\
{\modelname}$_{3b}$                            & \textbf{25.42}                  & \textbf{56.11}                    & \textbf{33.91}                    & \textbf{51.85}                    & \textbf{51.86}                       & \textbf{45.75}                     & \textbf{0.5789}    
\\ \midrule
\end{tabular}
\caption{Evaluation of differently-sized {\modelname} models on SQuAD. Base, Large and 3B refer to model configurations with 220 million, 770 million and 3 billion parameters, respectively. } 
\label{table:scaled}
\end{table*}

\end{document}